\documentclass[sigconf]{acmart} 
\usepackage[most]{tcolorbox}
\usepackage{listings}
\usepackage{caption}
\usepackage{natbib}  % DO NOT CHANGE THIS AND DO NOT ADD ANY OPTIONS TO IT
\usepackage{caption} % DO NOT CHANGE THIS AND DO NOT ADD ANY OPTIONS TO IT
\usepackage{algorithm}
\usepackage{algorithmic}
 \usepackage{amsmath}
 \usepackage{makecell}
 \usepackage{multirow}
\usepackage{booktabs}
\usepackage{amsmath}
\usepackage{tabularx}
\usepackage[most]{tcolorbox}
\usepackage{xcolor}
\usepackage{newfloat}
\usepackage{listings}
\AtBeginDocument{%
  }

\setcopyright{none}
\begin{document}

\title{ETAMem: Decoupling and Aligning Heterogeneous Memory for Long-Term Conversational Agents}


\author{Xinying Qian}
\email{qianxinying@dbis.nankai.edu.cn}
\affiliation{%
  \institution{Nankai University}
  \city{Tianjin}
  \country{China}
}

\author{Ying Zhang}
\email{yingzhang@nankai.edu.cn}
\affiliation{%
  \institution{Nankai University}
  \city{Tianjin}
  \country{China}
}

\author{Yu Zhao}
\email{zhaoyu@dbis.nankai.edu.cn}
\affiliation{%
  \institution{Nankai University}
  \city{Tianjin}
  \country{China}
}

\author{Xubo Liu}
\email{liuxubo@dbis.nankai.edu.cn}
\affiliation{%
  \institution{Nankai University}
  \city{Tianjin}
  \country{China}
}

\author{Baohang Zhou}
\email{zhoubaohang@dbis.nankai.edu.cn}
\affiliation{%
  \institution{Tiangong University}
  \city{Tianjin}
  \country{China}
}

\author{Jeff Z. Pan}
\email{j.z.pan@ed.ac.uk}
\affiliation{%
  \institution{The University of Edinburgh}
  \city{Edinburgh}
  \country{United Kingdom}
}

\renewcommand{\shortauthors}{Qian et al.}

\lstdefinestyle{promptstyle}{
    basicstyle=\ttfamily\scriptsize,
    breaklines=true,
    breakatwhitespace=true,
    columns=fullflexible,
    keepspaces=true,
    showstringspaces=false,
    frame=single,
    aboveskip=4pt,
    belowskip=4pt
}

\definecolor{promptdark}{HTML}{808080}
\definecolor{promptlight}{HTML}{F2F2F2}

\newtcolorbox{promptbox}[1]{
    enhanced,
    breakable,
    colback=promptlight,
    colframe=promptdark,
    colbacktitle=promptdark,
    coltitle=white,
    title={Prompt: #1},
    fonttitle=\bfseries,
    fontupper=\footnotesize,
    boxrule=0.8pt,
    arc=2mm,
    left=5pt,
    right=5pt,
    top=4pt,
    bottom=4pt,
    toptitle=3pt,
    bottomtitle=3pt,
    before skip=4pt,
    after skip=2pt,
    before upper={
        \setlength{\parindent}{0pt}
        \setlength{\parskip}{2pt}
    }
}

\newcommand{\promptvar}[1]{%
    \texttt{\detokenize{#1}}%
}

\begin{abstract}
Long-term memory enables Large Language Models (LLMs)
to maintain coherent and personalized interactions
across extended conversations.
However, existing memory systems face two key
challenges:
(1) \textit{Information Entanglement}, where
heterogeneous information is mixed within
coarse-grained memory units;
and (2) \textit{Organizational Mismatch}, where
memory structures based primarily on semantic
similarity may overlook other associations.
To address these challenges, we propose ETAMem,
a multi-view memory framework that decouples
heterogeneous conversational information and
aligns memory representations and organization
with its distinct characteristics.
Specifically, heterogeneous information parsing
extracts atomic memory units and identifies
affective information.
View-specific memory structuring then organizes
these units into three complementary hierarchical
views: Event, Affect, and Time.
ETAMem incrementally maintains memory structures
with adaptively calibrated thresholds.
For retrieval, entropy-based retrieval
allocates query-specific budgets across views
and performs hierarchical search within each view.
Experiments on LoCoMo and LongMemEval-S demonstrate
that ETAMem outperforms state-of-the-art baselines,
achieving relative accuracy improvements of
12.46\% and 7.96\%, respectively, while
maintaining competitive token efficiency.
Further analyses demonstrate the complementary
benefits of different memory views.
\end{abstract}




\received{20 February 2007}
\received[revised]{12 March 2009}
\received[accepted]{5 June 2009}

\maketitle

\section{Introduction}

Personalized conversational agents are increasingly
deployed in Web-based services~\cite{deng2025proactive},
where maintaining user context across interactions is essential for
coherent and personalized experiences~\cite{secom}.
To support long-term user modeling, memory systems
enable Large Language Models (LLMs) to retain and
retrieve information from conversations,
capturing users' experiences and preferences~\cite{du2025rethinking}.

Existing memory architectures for LLMs can be broadly
categorized into flat and structured approaches.
Flat approaches~\cite{memorybank,mem0} store extracted
conversational information as separate textual memory
entries, typically retrieved through embedding-based
similarity search.
In contrast, structured approaches~\cite{amem,timem,zep,hmem}
organize memories through hierarchies or graphs
to capture dependencies among memory units.
However, the heterogeneous nature of conversational
information poses two key challenges for
memory organization and retrieval.

\begin{figure}[t]
    \centering
    \includegraphics[width=\linewidth]{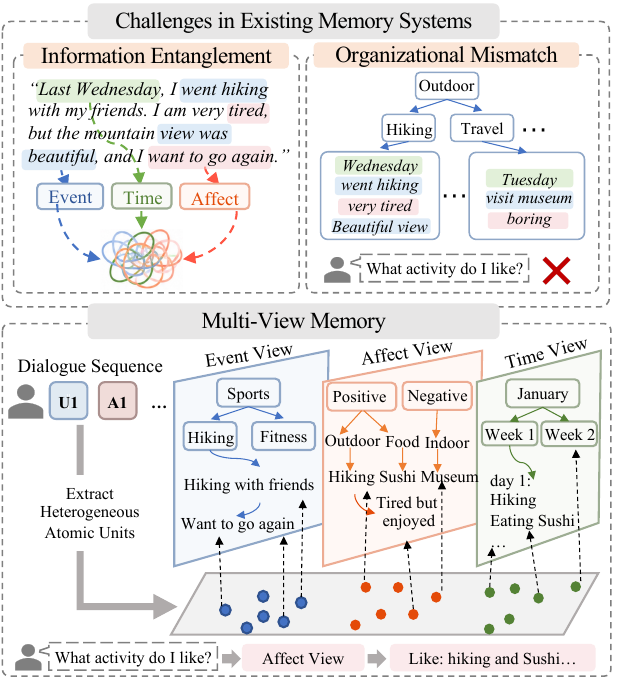}
\caption{Illustration of challenges in existing memory systems
and our proposed multi-view memory framework.}
    \label{fig:intro}
\end{figure}

\textbf{(1) Information Entanglement.}
Existing memory systems~\cite{memorybank,mem0}
often treat conversational memory entries as
the basic units of memory representation
and organization.
Although some approaches~\cite{amem,lightmem}
extract keywords or summaries, they do not
explicitly distinguish heterogeneous
information types.
As illustrated in Figure~\ref{fig:intro},
a single dialogue turn may contain factual
events, personal preferences, and temporal
contexts.
Without decomposition, these aspects remain
entangled, making it difficult to isolate
relevant evidence and introducing retrieval noise.

\textbf{(2) Organizational Mismatch.}
Existing memory systems~\cite{timem,zep}
organize memories through temporal or
relational structures.
Recent approaches~\cite{hmem,magma} further
combine multiple structures.
However, these structures do not necessarily
align memory representations and organization
criteria with different information types.
As shown in Figure~\ref{fig:intro},
organizing memories by semantic topics may
scatter related preferences across different
branches, making affective evidence difficult
to retrieve collectively.
%
% Similarly, temporal evidence may be dispersed
% across semantically distinct branches,
% hindering time-related retrieval.

These limitations motivate a fundamental question:
\textit{how can conversational
information be identified, represented,
and organized for
long-term memory retrieval?}
Different information types require distinct
representations and organizational principles.
For example, answering ``what activity do I like?''
requires connecting affective information
across different contexts, whereas
``what did I do this week?'' requires
aggregating experiences according to their
temporal relations.
This distinction aligns with findings from
cognitive neuroscience, which suggest that
episodic memory and emotional processing
involve functionally specialized yet
interacting neural
systems~\cite{phelps2004human,mcgaugh2004amygdala}.
This motivates decoupling heterogeneous
conversational information and aligning memory
representations and organization with its
distinct characteristics.

In this paper, we propose ETAMem, an
\textbf{E}vent-\textbf{T}ime-\textbf{A}ffect
\textbf{Mem}ory framework for long-term conversational
agents.
ETAMem first performs heterogeneous information
parsing by decomposing dialogue turns into
semantically complete atomic memory units (AMUs)
and identifying their affective characteristics.
Through view-specific memory structuring,
these AMUs are then organized into three hierarchical views.
Specifically, the Event view organizes factual
experiences through semantic similarity,
the Affect view organizes emotions, attitudes,
and preferences through affect-aware representations,
and the Time view indexes all memories through temporal hierarchies.
To support continual interaction, ETAMem
incrementally maintains the Event and Affect
hierarchies through attachment, splitting, merging,
and lazy summary updates, while adaptively calibrating
maintenance thresholds.
Given a query, entropy-based  retrieval
allocates budgets across views
through adaptive routing and performs hierarchical
beam search to retrieve relevant memories.
Experiments on two long-term dialogue memory
benchmarks demonstrate that ETAMem consistently
improves memory performance, particularly on
multi-hop and temporally grounded queries.
Overall, our work makes the following contributions:
\begin{itemize}
    
\item We propose ETAMem, a multi-view memory
framework that organizes heterogeneous
conversational information into complementary
Event, Affect, and Time hierarchies with
view-specific representations.

    \item We develop a dynamic memory maintenance
    mechanism with adaptive threshold calibration.
    We further employ entropy-based routing to
    allocate query-specific retrieval budgets
    across complementary memory views.

    \item Extensive experiments on LoCoMo and
    LongMemEval-S demonstrate that ETAMem outperforms
    strong baselines, particularly on multi-hop and
    temporal reasoning questions, while maintaining
    competitive token costs.
\end{itemize}

\section{Related Work}

\subsection{Memory Systems for LLM Agents}

\textbf{Flat memory.}
Early memory systems store extracted facts or interaction
summaries as independent units, typically retrieved through
semantic similarity~\cite{memorybank,mem0}.
For long-term access, MemGPT~\cite{memgpt} separates
in-context and external memory through paging,
MemoryOS~\cite{memoryos} organizes histories into
short-, mid-, and long-term layers, and
LightMem~\cite{lightmem} uses staged offline consolidation.
These designs emphasize where and when memories are
stored or consolidated, while retrieval often relies
on general semantic matching.

\textbf{Structured memory.}
Structured systems organize memories into hierarchies
or graphs.
Tree-based approaches support memory abstraction
at different levels of granularity.
TiMem~\cite{timem} builds a five-level temporal memory
tree to consolidate experiences over time,
while xMemory~\cite{hu2026xmemory} decomposes interaction
histories into semantic components and organizes
them hierarchically.
Graph-based approaches emphasize explicit
memory associations.
A-MEM~\cite{amem} dynamically updates semantic
links between memory notes, whereas Zep~\cite{zep}
maintains a temporal knowledge graph of evolving
facts and relationships.
HingeMem~\cite{hingemem} uses boundary-based
hypergraphs to connect memories by person,
time, location, and topic.
However, these methods primarily emphasize
temporal, semantic, or relational organization
without explicitly aligning memory hierarchies
with different types of conversational information.

\textbf{Multi-structure memory.}
Recent work organizes memory through multiple
complementary structures or information types.
MemWeaver~\cite{memweaver} develops complementary
behavioral and cognitive memory hierarchies
to capture user interactions and long-term preferences.
H-Mem~\cite{hmem} combines a temporal-semantic tree
with an entity knowledge graph.
MAGMA~\cite{magma} links memory events through separate
semantic, temporal, causal, and entity graphs.
MemGAS~\cite{xu2026memgas} associates multi-granularity
memory units through clustering and selects retrieval
granularity via entropy-based routing.
MemoType~\cite{wen2026memotype} distinguishes
different memory types without hierarchical
organization.
These methods either combine multiple
memory structures without explicitly decoupling
heterogeneous information or distinguish
information types without organizing them
into specialized hierarchies.

\subsection{Affect-Aware Memory and Retrieval}

Recent studies have explored affective information
to improve memory retrieval and personalized interactions.
Emotional RAG~\cite{emotionrag} combines emotional
and semantic relevance to retrieve memories
for role-playing agents.
REG~\cite{wang2026reg} retrieves emotionally
similar examples to support empathetic
response generation.
PAMU~\cite{sun2026preference} dynamically updates
memories to reflect evolving user preferences,
while Sentipolis~\cite{fu-etal-2026-sentipolis}
models emotion dynamics and emotion-memory
coupling to maintain affective continuity.
Recent benchmarks further emphasize the importance
of affective and personalized memory.
CloneMem~\cite{hu-etal-2026-clonemem}
evaluates the retention and retrieval of
personal experiences, preferences, and opinions,
while ENPMR-Bench~\cite{fu2026enpmr}
focuses on memory retrieval for emotional support.
Despite these advances, existing approaches
primarily leverage affective information to
guide retrieval, memory updates, or generation,
with limited attention to organizing memories
through affective associations.
\begin{figure*}[t]
    \centering
    \includegraphics[width=\textwidth]{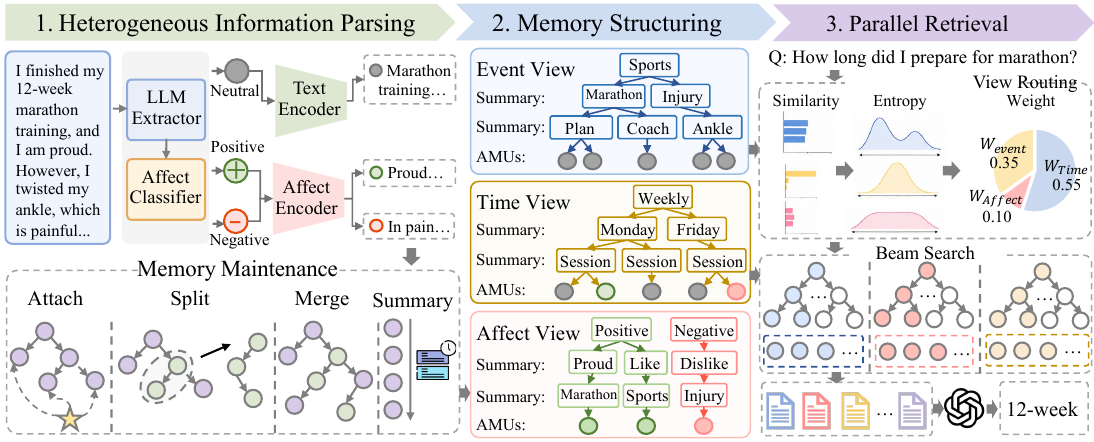}
\caption{Overview of {ETAMem}. Given dialogue histories, the framework extracts AMUs and organizes them into three memory views: Event, Affect, and Time. During retrieval, entropy-based routing and beam search are used for response generation.}
    \label{fig:method}
\end{figure*}
\section{Method}
\label{sec:method}

We present ETAMem, a multi-view memory framework
for decoupling and aligning heterogeneous
conversational information.
First, heterogeneous information parsing extracts
atomic memory units (AMUs) and identifies their
affective characteristics.
These units are then organized through view-specific
memory structuring into three complementary views:
Event, Affect, and Time.
To support evolving conversations, dynamic memory
maintenance updates the hierarchies, while adaptive
calibration adjusts maintenance thresholds.
Finally, entropy-based  retrieval allocates
query-specific budgets across views and performs
hierarchical beam search to retrieve relevant AMUs,
as shown in Figure~\ref{fig:method}.

% Notation and algorithms are provided in
% Appendix~\ref{method_details}.

\subsection{Problem Formulation}
\label{sec:problem}

The long-term memory requires LLMs to continuously organize past interactions and retrieve relevant information for reasoning.
Let $\mathcal{D}_T=\{(d_t,\tau_t)\}_{t=1}^{T}$ denote a dialogue history processed sequentially, where $d_t$ and $\tau_t$ represent the dialogue content and timestamp at turn $t$, respectively.
At each turn $t$, a memory state $\mathcal{M}_t$ is incrementally updated from the previous state $\mathcal{M}_{t-1}$ and the new dialogue input $(d_t,\tau_t)$, where $\mathcal{M}_t$ denotes the organized memory after processing $t$ turns.
Given a query $q$, the objective is to retrieve an evidence set $\mathcal{E}_q$ from $\mathcal{M}_T$ that supports accurate inference.

\subsection{Heterogeneous Information Parsing}
\label{sec:amu}

\textbf{AMU Extraction.}
ETAMem decomposes each dialogue turn into atomic memory units (AMUs), such that each AMU captures one independent memory item and the resulting set preserves the original information. It then assigns a polarity label to each AMU to distinguish factual content from affective content for subsequent memory organization.

Given a dialogue turn $x_t=(d_t,\tau_t)$, ETAMem uses an LLM to extract a set of AMU contents:
\begin{equation}
    \mathcal{A}_t
    = f_{\mathrm{LLM}}(x_t)
    = \{a_{t,1},\ldots,a_{t,n_t}\}.
\end{equation}
Here, $n_t$ is the number of AMUs extracted from turn $t$.

To guide view-specific memory organization, ETAMem assigns each AMU a polarity label. We define the polarity label space as
$\mathcal{P}=\{\mathrm{negative},\mathrm{neutral},\mathrm{positive}\}$.
Given the class scores produced by an off-the-shelf sentiment classifier $f_{\mathrm{sent}}$, the highest-scoring class is selected as
\begin{equation}
    p_{t,i}
    =
    \arg\max_{\ell\in\mathcal{P}}
    \left[f_{\mathrm{sent}}(a_{t,i})\right]_\ell .
\end{equation}

% \textbf{AMU Representation.}
% We use two encoders to capture semantic and affective information, where
% $\phi_{\mathrm{text}}(\cdot)$ denotes the textual encoder and
% $\phi_{\mathrm{affect}}(\cdot)$ denotes the affective encoder.
% Each AMU is encoded according to its polarity label:
% \begin{equation}
% \mathbf r_{t,i}
% =
% \begin{cases}
% \phi_{\mathrm{text}}(a_{t,i}),
% & p_{t,i}=\mathrm{neutral},\\
% \phi_{\mathrm{affect}}(a_{t,i}),
% & p_{t,i}\in\{\mathrm{negative},\mathrm{positive}\}.
% \end{cases}
% \end{equation}

\subsection{View-Specific Memory Structuring}
\label{sec:tri_tree}

\textbf{AMU Representation.}
We use two encoders to capture semantic and
affective information, where
$\phi_{\mathrm{text}}(\cdot)$ denotes the textual
encoder and $\phi_{\mathrm{affect}}(\cdot)$ denotes
the affective encoder.
Each AMU is encoded according to its polarity label:
\begin{equation}
\mathbf r_{t,i}
=
\begin{cases}
\phi_{\mathrm{text}}(a_{t,i}),
& p_{t,i}=\mathrm{neutral},\\
\phi_{\mathrm{affect}}(a_{t,i}),
& p_{t,i}\in\{\mathrm{negative},\mathrm{positive}\}.
\end{cases}
\end{equation}

\textbf{Memory Organization.}
To support structured organization and
coarse-to-fine retrieval over heterogeneous
AMUs, ETAMem organizes the current memory
as three hierarchical views:
\begin{equation}
\mathcal{M} =
\left\{
\mathcal{V}_{\mathrm{event}},
\mathcal{V}_{\mathrm{affect}},
\mathcal{V}_{\mathrm{time}}
\right\}.
\end{equation}

\textbf{The Event View}
$\mathcal{V}_{\mathrm{event}}$ organizes neutral
AMUs according to textual semantics.
AMUs with similar semantic representations
are hierarchically grouped into event-related
nodes, capturing recurring events, entities,
and interactions.

\textbf{The Affect View}
$\mathcal{V}_{\mathrm{affect}}$ organizes positive
and negative AMUs according to affective semantics.
AMUs with similar affective representations
are hierarchically grouped into nodes,
capturing persistent preferences and attitudes.

\textbf{The Time View}
$\mathcal{V}_{\mathrm{time}}$ organizes AMUs
through a temporal hierarchy:
$\mathrm{Session}\rightarrow\mathrm{Day}
\rightarrow\mathrm{Week}\rightarrow\mathrm{Month}$.
AMUs are stored at the Session level,
while higher-level nodes provide summaries.
Unlike Event and Affect views, the Time view
follows predefined temporal granularity
rather than semantic grouping.

\textbf{Tree Structure.}
Each memory view adopts a hierarchical structure.
Leaf nodes store AMUs, while internal nodes
maintain textual summaries of their subtrees.
For an internal node whose children are leaves,
its summary is generated from the AMUs stored
in those leaves.
For higher-level internal nodes, the summary
is generated from the summaries of their
child nodes.

\subsection{Dynamic Memory Maintenance}
\label{sec:online_update}
\subsubsection{Memory Maintenance}
ETAMem maintains its memory trees through online insertion and incremental structural refinement. For each incoming dialogue turn, the extracted AMUs are inserted into their corresponding views. At the end of each session, the Event and Affect views undergo structural maintenance, with Split and Merge operations triggered only when their respective conditions are met. In contrast, the Time view follows a predefined temporal hierarchy and requires only deterministic insertion without maintenance.

\textbf{Attachment.}
The attachment operation determines how a new AMU is incorporated into each memory view.
For the Time view, an AMU is assigned to the temporal node corresponding to its timestamp.
For the Event and Affect views, ETAMem identifies the most suitable insertion position through hierarchical search.

We first represent each node by the centroid of the AMU representations in its subtree:
\begin{equation}
\boldsymbol{\mu}_n
=
\frac{1}{|S_n|}
\sum_{a_{t,i}\in S_n}\mathbf r_{t,i},
\end{equation}
where $S_n$ denotes the set of AMUs contained in the subtree rooted at $n$.
The similarity between an AMU and a candidate node is measured as
$
g(a_{t,i},n)
=
\cos(\mathbf r_{t,i},\boldsymbol{\mu}_n).
$

Using this similarity score, ETAMem performs beam search over corresponding trees.
Starting from the root nodes, it recursively expands child nodes and retains the top-$B$ candidate paths at each layer until reaching the leaf level, forming a candidate set $\mathcal{L}_B$.
The insertion candidate is selected as
$
n^\star=\arg\max_{n\in\mathcal{L}_B}g(a_{t,i},n),
$
with score $\rho^\star=g(a_{t,i},n^\star)$.

Based on the matching score, ETAMem selects one of three insertion operations for the AMU: \textsc{Attach} to the most relevant existing leaf, \textsc{NewLeaf} under the parent of $n^\star$, or \textsc{NewRoot}.
{\small
\begin{equation}
\small
\Gamma(a_{t,i},n^{\star})=
\begin{cases}
\mathrm{Attach}(a_{t,i},n^\star),
& \rho^\star\geq\theta_{\mathrm{attach}},\\
\mathrm{NewLeaf}(a_{t,i},n^\star),
& \theta_{\mathrm{new}}\leq\rho^\star<\theta_{\mathrm{attach}},\\
\mathrm{NewRoot}(a_{t,i}),
& \rho^\star<\theta_{\mathrm{new}}.
\end{cases}
\end{equation}
}

\textbf{Split.}
To prevent leaf nodes from becoming overly large or semantically incoherent, ETAMem splits a leaf when its size exceeds \(M_{\max}\) or its cohesion falls below \(\theta_{\mathrm{coh}}\).
For a leaf containing \(m\) AMUs, ETAMem partitions the AMUs using cosine K-means, with the number of clusters determined as
$
k
=
\max\left(
2,
\left\lceil\frac{m}{M_{\max}}\right\rceil
\right).
$
Each cluster forms a new leaf under the same parent.
The cohesion of a leaf node \(n\) is computed as
\begin{equation}
\small
\kappa(n)=
\frac{2}{m(m-1)}
\sum_{j<\ell}
\cos\left(
\mathbf r_{t_j,i_j},
\mathbf r_{t_\ell,i_\ell}
\right).
\end{equation}

% \textbf{Merge.}
% To reduce redundant small nodes and improve retrieval efficiency, ETAMem checks the Event and Affect views for merging at the end of each dialogue session. For each candidate pair of nodes $n_i$ and $n_j$, their similarity is computed from the node centroids as $\mu_{ij}=\cos(\boldsymbol{\mu}_{n_i},\boldsymbol{\mu}_{n_j})$. The pair is merged when $\mu_{ij}\geq\theta_{\mathrm{merge}}$. For leaf nodes, the merged AMU set must additionally satisfy $|S_{n_i}\cup S_{n_j}|\leq M_{\max}$. When the two nodes belong to different branches, the merged node is placed under their nearest common ancestor to preserve the existing hierarchy.

\textbf{Merge.}
To reduce redundant small nodes and improve retrieval efficiency, ETAMem checks the Event and Affect views for merging at the end of each dialogue session. For each candidate pair of nodes $n_i$ and $n_j$, their similarity is computed from the node centroids as $s_{\mathrm{merge}}(n_i,n_j)=\cos\left(\boldsymbol{\mu}_{n_i},\boldsymbol{\mu}_{n_j}\right)$. The pair is merged when $s_{\mathrm{merge}}(n_i,n_j)\geq\theta_{\mathrm{merge}}$. For leaf nodes, the merged AMU set must additionally satisfy $|S_{n_i}\cup S_{n_j}|\leq M_{\max}$. When the two nodes belong to different branches, the merged node is placed under their nearest common ancestor to preserve the existing hierarchy.

\textbf{Summary Update.}
As memory trees evolve, frequent updates to internal node summaries would introduce additional LLM overhead.
To reduce recomputation, ETAMem lazily updates internal summaries using a pending-change counter $u_n$.
Each structural update increments the counters of affected nodes and their ancestors; when $u_n\geq\delta$, where $\delta$ is the update interval controlling how many pending updates trigger summary regeneration, the summary $z_n$ is regenerated and $u_n$ is reset.
Leaf summaries are directly updated from their stored AMU texts.

\subsubsection{Adaptive Calibration}
\label{sec:adaptive_calibration}

To prevent excessive tree growth and node
overcrowding as memory expands, ETAMem
adaptively calibrates structural parameters
and maintenance thresholds.
Both are initialized with predefined values
and subsequently adjusted using observed
memory statistics, without reconstructing
existing trees.

\textbf{Structural Parameters.}
For each view $v$, ETAMem calibrates the maximum
AMUs per leaf $M_{\max}^{(v)}$, maximum children
per internal node $b_{\max}^{(v)}$, and maximum
tree depth $H^{(v)}$.
When preset capacity limits are reached,
these parameters are recalibrated to accommodate
further memory growth.
Let $\widetilde{n}_v$ and $n_{\max}^{(v)}$
denote the median and maximum AMU counts
across observed dialogue histories in view $v$.
The parameters are computed as
\begin{equation}
\begin{aligned}
\small
M_{\max}^{(v)}
&=\max\bigl(2,\operatorname{round}
(\sqrt{\widetilde{n}_v})\bigr),\\
b_{\max}^{(v)}
&=\max\bigl(2,\lceil
\sqrt{\widetilde{n}_v/M_{\max}^{(v)}}
\rceil\bigr),\\
H^{(v)}
&=\max\bigl(2,1+
\lceil\log_{b_{\max}^{(v)}}{\widehat{L}_v}\rceil\bigr),
\end{aligned}
\label{eq:structural_calibration}
\end{equation}
where $\widehat{L}_v=\max(1,\lceil
n_{\max}^{(v)}/M_{\max}^{(v)}
\rceil)$ estimates the required leaves.
This allows the tree structure to adapt
to increasing memory size.

\textbf{Maintenance Thresholds.}
ETAMem calibrates view-specific thresholds
based on observed similarity distributions.
During memory insertion, cosine similarities
between incoming AMUs and visited candidate
nodes are collected across dialogue histories.
We employ Otsu's method~\cite{otsu1979threshold}
to partition these similarities into low-
and high-similarity subsets.
Specifically, $\theta_{\mathrm{new}}$ is estimated
as the 75th percentile of the low-similarity subset,
while $\theta_{\mathrm{attach}}$,
$\theta_{\mathrm{coh}}$, and
$\theta_{\mathrm{merge}}$ are estimated
as the 25th, 65th, and 85th percentiles
of the high-similarity subset.

To reduce fluctuations across calibration rounds,
ETAMem updates each threshold using an
exponential moving average
(EMA)~\cite{roberts1959control}:
\begin{equation}
\small
\theta_c^{(v,s)}
=
\eta\theta_c^{(v,s-1)}
+
(1-\eta)\widehat{\theta}_c^{(v,s)},
\label{eq:adaptive_calibration}
\end{equation}
where $s$ indexes calibration rounds,
$c\in\{\mathrm{new},\mathrm{attach},
\mathrm{coh},\mathrm{merge}\}$ denotes
the threshold type,
$\widehat{\theta}_c^{(v,s)}$ is its current
estimate for view $v$, and $\eta$ controls
the contribution of the previous threshold.

\subsection{Entropy-based  Retrieval}
\label{sec:retrieval}

\textbf{Entropy-based View Routing.}
Different queries may depend on different memory views and thus require different types of evidence. To address this, ETAMem computes the confidence of each view and adaptively allocates the retrieval budget across the three views.

Let \(\mathcal{V}=\{\mathrm{Event},\mathrm{Affect},\mathrm{Time}\}\) denote the set of memory views.
For each view $v\in\mathcal{V}$, let $\mathcal{L}_v$ denote its leaf set,
$N_v=|\mathcal{L}_v|$ the number of leaves, and $z_{v,j}$ the summary of its
$j$-th leaf. We first compute the similarity between the query $q$ and each leaf summary:
\begin{equation}
s_{v,j}
=
\cos\left(
\phi_{\mathrm{text}}(q),
\phi_{\mathrm{text}}(z_{v,j})
\right),
\quad j=1,\ldots,N_v.
\end{equation}

The similarity scores are converted into a probability distribution over the leaves within each view. $\beta$ controls the temperature.
\begin{equation}
\small
\alpha_{v,j}
=
\frac{\exp(s_{v,j}/\beta)}
{\sum_{\ell=1}^{N_v}\exp(s_{v,\ell}/\beta)},
\end{equation}
We then use normalized entropy to measure how concentrated the probability distribution is:
\begin{equation}
\small
\bar{H}_v
=
-\frac{1}{\log\max\{2,|\mathcal{L}_v|\}}
\sum_{j=1}^{N_v}
\alpha_{v,j}\log\alpha_{v,j}.
\end{equation}
A lower $\bar{H}_v$ indicates that the query is concentrated on fewer leaves, suggesting a clearer retrieval signal for view $v$.
We use $1-\bar{H}_v$ as the routing score, where larger values indicate more concentrated retrieval distributions, and normalize it across views to obtain the routing weight $w_v$:
\begin{equation}
\small
w_v
=
\frac{1-\bar{H}_v}
{\sum_{u\in\mathcal{V}}(1-\bar{H}_u)}.
\end{equation}

Given a total leaf retrieval budget $K$, ETAMem initializes the per-view
budget as $K_v=\max(1,\operatorname{round}(Kw_v))$.
Thus, views with more concentrated leaf-level similarity distributions receive larger shares of the retrieval budget.

\textbf{ Beam Search.}
For each memory view, ETAMem retrieves relevant nodes through layer-wise beam search by combining semantic similarity and lexical relevance, where semantic similarity captures contextual relevance while lexical relevance preserves exact term matching.
The summary of candidate node $n$ is denoted as $z_n$, and its semantic relevance with query $q$ is computed as
\begin{equation}
    s_{\mathrm{sem}}(q,n)
    =
    \cos\left(
        \phi_{\mathrm{text}}(q),
        \phi_{\mathrm{text}}(z_n)
    \right).
\end{equation}

The lexical relevance is measured by BM25~\cite{bm25}.
The normalized semantic and lexical scores are combined as
\begin{equation}
    \mathrm{score}(q,n)
    =
    (1-\lambda)\widetilde{s}_{\mathrm{sem}}(q,n)
    +
    \lambda\widetilde{s}_{\mathrm{lex}}(q,n),
\end{equation}
where $\widetilde{s}$ denotes the normalized score and $\lambda$ controls the contribution of lexical matching.

Starting from the root nodes, ETAMem retains
the top-$B$ nodes according to $\mathrm{score}(q,n)$
at each internal layer and recursively expands them.
After expanding the final internal-layer beam,
it selects the top-$K_v$ leaf nodes for view $v$.
The retrieved AMUs across views are merged
and deduplicated to form the evidence set
$\mathcal{E}_q$.
If the LLM returns ``no information'' due to insufficient evidence, ETAMem falls back
to the corresponding original dialogue turns
$\mathcal{D}_q$.
The response is generated as
\begin{equation}
\hat{y} =
\begin{cases}
f_{\mathrm{LLM}}(q,\mathcal{E}_q),
& \text{if sufficient},\\
f_{\mathrm{LLM}}(q,\mathcal{D}_q),
& \text{otherwise}.
\end{cases}
\end{equation}

\section{Experiments}
\begin{table}[t]
\caption{Final parameter configurations of ETAMem.}
\label{tab:calibration_settings}
\centering
\small
\setlength{\tabcolsep}{4pt}
\renewcommand{\arraystretch}{1.08}

\resizebox{0.9\columnwidth}{!}{
\begin{tabular}{llccccccc}
\toprule
\multirow{2}{*}{Dataset}
& \multirow{2}{*}{View}
& \multicolumn{3}{c}{Structural Parameters}
& \multicolumn{4}{c}{Maintenance Thresholds} \\
\cmidrule(lr){3-5}
\cmidrule(l){6-9}
&
& \(M_{\max}^{(v)}\)
& \(b_{\max}^{(v)}\)
& \(H^{(v)}\)
& \(\theta_{\mathrm{new}}\)
& \(\theta_{\mathrm{attach}}\)
& \(\theta_{\mathrm{coh}}\)
& \(\theta_{\mathrm{merge}}\) \\
\midrule

\multirow{2}{*}{LoCoMo}
& Event  & 29 & 6 & 4 & 0.402 & 0.460 & 0.515 & 0.572 \\
& Affect & 33 & 6 & 4 & 0.446 & 0.504 & 0.562 & 0.619 \\

\midrule

\multirow{2}{*}{LongMemEval}
& Event  & 22 & 5 & 4 & 0.426 & 0.481 & 0.534 & 0.582 \\
& Affect & 11 & 4 & 4 & 0.435 & 0.486 & 0.537 & 0.582 \\

\bottomrule
\end{tabular}
}

\end{table}
\subsection{Experimental Setup}

\textbf{Datasets.}
We conduct experiments on two long-term memory benchmarks. LoCoMo~\cite{locomo} contains multi-session interactions from 10 user groups, while LongMemEval-S~\cite{longmemeval} comprises 500 questions designed to assess memory over extended contexts.
Further details of the datasets are provided
in Appendix~\ref{sec:dataset_details}.

\textbf{Baselines.} We compare ETAMem with representative memory baselines:
MemoryBank~\cite{memorybank},
A-MEM~\cite{amem},
Mem0~\cite{mem0},
MemoryOS~\cite{memoryos},
MemOS~\cite{memos},
MAGMA~\cite{magma},
TiMem~\cite{timem},
LightMem~\cite{lightmem}.

\begin{table*}[t]
\caption{Performance and token costs on LoCoMo.
Build and Inference denote token costs for memory
construction and inference, averaged per session (k)
and per question, respectively.}
\label{locomo}
\centering
\small
\begin{tabular}{lccccccc}
\toprule
\multirow{2}{*}{\textbf{Method}}
& \textbf{Single-Hop}
& \textbf{Temporal}
& \textbf{Open-Domain}
& \textbf{Multi-Hop}
& \textbf{Overall}
& \multicolumn{2}{c}{\textbf{Token Cost}} \\
\cmidrule(lr){7-8}
&
$\uparrow$ (841Q)
&
$\uparrow$ (321Q)
&
$\uparrow$ (96Q)
&
$\uparrow$ (282Q)
&
$\uparrow$ (1540Q)
&
\textbf{Build (k)}
&
\textbf{Inference} \\
\midrule

MemoryBank
& $46.18 \pm 0.32$
& $29.34 \pm 0.45$
& $36.67 \pm 0.47$
& $33.36 \pm 0.54$
& $39.77 \pm 0.27$
& -- & 8063.77 \\

A-MEM
& $52.82 \pm 0.28$
& $60.87 \pm 0.37$
& $43.75 \pm 0.00$
& $38.37 \pm 0.27$
& $51.29 \pm 0.06$
& 1149.43 & 2431.40 \\

Mem0
& $62.09 \pm 0.42$
& $59.25 \pm 0.41$
& $37.70 \pm 0.47$
& $50.14 \pm 0.32$
& $57.79 \pm 0.34$
& 1693.39 & \textbf{1070.10} \\

MemoryOS
& $68.37 \pm 0.46$
& $52.46 \pm 0.49$
& $46.67 \pm 1.79$
& $52.76 \pm 0.49$
& $60.79 \pm 0.48$
& 287.00 & 4659.09 \\

MemOS
& $76.07 \pm 0.10$
& $69.47 \pm 0.25$
& $45.14 \pm 0.49$
& $56.85 \pm 0.69$
& $69.24 \pm 0.11$
& -- & 1371.42 \\

MAGMA
& $81.33 \pm 0.78$
& $59.81 \pm 0.54$
& \underline{$58.33 \pm 1.04$}
& \underline{$69.50 \pm 0.71$}
& $73.25 \pm 0.62$
& 434.93 & 2055.39 \\

TiMem
& \underline{$81.43 \pm 0.05$}
& $\underline{77.63 \pm 0.34}$
& $52.08 \pm 0.74$
& $62.20 \pm 0.82$
& \underline{$75.30 \pm 0.16$}
& 411.66 & 7627.67 \\

LightMem
& $74.20 \pm 0.43$
& $70.09 \pm 0.62$
& $51.04 \pm 0.00$
& $57.80 \pm 0.35$
& $68.90 \pm 0.34$
& \textbf{80.48}
& \underline{1206.32} \\

\midrule

\textbf{ETAMem}
& $\boldsymbol{88.47 \pm 0.59}$
& $\boldsymbol{80.06 \pm 0.82}$
& $\boldsymbol{80.21 \pm 1.80}$
& ${80.14 \pm 0.61}$
& $\boldsymbol{84.68 \pm 0.53}$
& \underline{132.06}
& 1331.60 \\

% Qwen accuracy from supplied experimental logs
% Token costs retained from original table; verify.

w/ Qwen3-30B
& $88.59 \pm 1.61$
& $63.86 \pm 0.93$
& $68.75 \pm 2.08$
& $\boldsymbol{81.21 \pm 0.61}$
& $80.84 \pm 1.10$
& 139.76
& 1368.63 \\

\bottomrule
\end{tabular}
\end{table*}
\begin{table*}[t]
\caption{Performance and token costs on LongMemEval-S.
KU: Knowledge Update; MS: Multi-Session;
SSA/P/U: Single-Session Assistant/Preference/User;
TR: Temporal Reasoning.
Build and Inference denote token costs for memory construction
and inference, averaged per session (k) and per question, respectively.}
\label{longmemeval_results}
\centering
\small
\begin{tabular}{lccccccccc}
\toprule
\multirow{2}{*}{\textbf{Method}}
& \textbf{KU}
& \textbf{MS}
& \textbf{SSA}
& \textbf{SSP}
& \textbf{SSU}
& \textbf{TR}
& \textbf{Overall}
& \multicolumn{2}{c}{\textbf{Token Cost}} \\
\cmidrule(lr){9-10}
&
$\uparrow$ (78Q)
&
$\uparrow$ (133Q)
&
$\uparrow$ (56Q)
&
$\uparrow$ (30Q)
&
$\uparrow$ (70Q)
&
$\uparrow$ (133Q)
&
$\uparrow$ (500Q)
&
\textbf{Build (k)}
&
\textbf{Inference} \\
\midrule

MemoryBank
& $21.79 \pm 0.00$
& $9.77 \pm 0.00$
& $50.00 \pm 0.00$
& $12.00 \pm 1.83$
& $29.71 \pm 0.64$
& $17.14 \pm 0.34$
& $21.04 \pm 0.09$
& -- & 13906.81 \\

A-MEM
& $72.82 \pm 0.51$
& $40.30 \pm 0.37$
& $87.50 \pm 0.00$
& $39.33 \pm 2.49$
& $82.86 \pm 0.00$
& $36.09 \pm 0.48$
& $55.44 \pm 0.15$
& 1605.81 & 3971.60 \\

Mem0
& $78.72 \pm 0.70$
& $66.17 \pm 0.92$
& $51.79 \pm 0.00$
& $50.00 \pm 2.36$
& $94.29 \pm 0.00$
& $49.17 \pm 0.67$
& $64.96 \pm 0.41$
& 1152.62 & 1647.56 \\

MemoryOS
& $56.15 \pm 0.57$
& $44.81 \pm 0.41$
& $78.18 \pm 0.00$
& $51.33 \pm 1.83$
& $81.14 \pm 0.64$
& $53.38 \pm 0.00$
& $58.04 \pm 0.18$
& 2938.41 & 7574.30 \\

MemOS
& $76.67 \pm 0.51$
& $58.80 \pm 0.30$
& $67.86 \pm 0.00$
& $50.67 \pm 1.33$
& $93.71 \pm 0.70$
& $65.11 \pm 0.37$
& $68.68 \pm 0.16$
& -- & \underline{1091.51} \\

MAGMA
& $65.38 \pm 2.22$
& $34.59 \pm 1.30$
& $\boldsymbol{98.21 \pm 0.00}$
& $\underline{90.00 \pm 0.00}$
& $88.57 \pm 1.43$
& $25.56 \pm 1.30$
& $55.00 \pm 0.60$
& 196.34
& 5372.82 \\

TiMem
& $\underline{86.16 \pm 1.07}$
& $70.83 \pm 0.98$
& $82.14 \pm 0.00$
& $63.33 \pm 0.00$
& $\boldsymbol{95.71 \pm 0.00}$
& $68.42 \pm 0.00$
& $\underline{76.88 \pm 0.30}$
& 563.40 & 7164.91 \\

LightMem
& $85.90 \pm 1.28$
& $\underline{77.44 \pm 0.75}$
& $33.93 \pm 0.00$
& $73.33 \pm 0.00$
& $94.29 \pm 1.43$
& $\underline{71.43 \pm 0.75}$
& $74.40 \pm 0.53$
& \textbf{28.25}
& \textbf{749.95} \\

\midrule

\textbf{ETAMem}
& $\boldsymbol{87.18 \pm 1.28}$
& $\boldsymbol{78.95 \pm 0.75}$
& $\underline{92.86 \pm 0.00}$
& $76.67 \pm 0.00$
& $\boldsymbol{95.71 \pm 0.00}$
& $\boldsymbol{75.19 \pm 0.75}$
& $\boldsymbol{83.00 \pm 0.20}$
& 63.41
& 2343.64 \\

% Qwen accuracy from supplied experimental logs
% Token costs retained from original table; verify.
w/ Qwen3-30B
& $\boldsymbol{87.18 \pm 1.28}$
& $70.68 \pm 0.75$
& $91.07 \pm 0.00$
& $\boldsymbol{93.33 \pm 0.00}$
& $\boldsymbol{95.71 \pm 0.00}$
& $70.68 \pm 1.50$
& $\underline{80.40 \pm 0.20}$
& \underline{56.10}
& 2582.10 \\

\bottomrule
\end{tabular}
\end{table*}
\textbf{Implementation Details.}
For fair comparison, we follow the same
evaluation settings as TiMem.
Specifically, all methods use {gpt-4o-mini} as the LLM backbone and {Qwen3-Embedding-0.6B} as the textual embedding model.
We additionally evaluate ETAMem with
Qwen3-30B-A3B-Instruct-2507, denoted as
\textit{w/ Qwen3-30B}.
The retrieval budget is set to $K=20$ for all methods.
For ETAMem, we set $\lambda=0.25$, beam width $B=10$, $\delta=20$, and routing temperature $\beta=0.1$.
The remaining parameters are described in the calibration settings.
For the Affect view, we employ two RoBERTa-based affective models: Twitter-RoBERTa-Sentiment~\cite{tweeteval} as the classifier and RoBERTa-GoEmotions~\cite{goemotions} as the encoder.
ETAMem is evaluated over three runs, with results reported as mean accuracy $\pm$ sample standard deviation in Tables~\ref{locomo} and~\ref{longmemeval_results}.
We evaluate all methods using LLM-as-a-judge accuracy.
Following~\cite{lightmem}, memory construction cost is measured as the average number of total tokens consumed by LLM calls during memory construction per session, while inference cost is measured as the average number of input tokens consumed per query.
All prompts are provided in Appendix~\ref{sec:prompt_design}.

\textbf{Calibration Settings.}
ETAMem adopts shared initial settings
 {across datasets for the Event and Affect views}.
The leaf capacity $M_{\max}^{(v)}$,
branching factor $b_{\max}^{(v)}$,
and maximum tree depth $H^{(v)}$
are initialized to 25, 5, and 4, respectively.
The thresholds $\theta_{\mathrm{new}}$
and $\theta_{\mathrm{attach}}$
are initialized to 0.4 and 0.5,
while $\theta_{\mathrm{coh}}$
and $\theta_{\mathrm{merge}}$
are initialized to 0.6 and 0.7.
The calibration percentiles are set to
75, 25, 65, and 85, respectively,
with EMA coefficient ${\eta}=0.5$.
Structural parameters are recalibrated at
capacity limits, while maintenance thresholds
are updated after each session, with changes
applied to the next session.
Table~\ref{tab:calibration_settings}
reports the final calibrated parameter values.

\subsection{Main Results}

Table~\ref{locomo} presents the results on LoCoMo.
ETAMem achieves the highest overall accuracy of
84.68\%, outperforming the strongest memory baseline,
TiMem, by 12.46\%.
Compared with TiMem's temporal hierarchy and MAGMA's
multi-graph architecture, ETAMem improves overall
accuracy by 9.38 and 11.43 percentage points,
respectively.
These improvements suggest the benefits of
organizing heterogeneous memories through
complementary, view-specific structures.
Notably, ETAMem achieves relative improvements
of 15.31\% and 37.51\% on multi-hop and open-domain
questions over the strongest baselines, respectively.
In terms of efficiency, ETAMem reduces memory
construction tokens by 67.92\% and 88.51\%
compared with TiMem and A-MEM, respectively.

Table~\ref{longmemeval_results} reports the results
on LongMemEval-S.
ETAMem achieves the highest overall accuracy of
83.00\%, surpassing TiMem by 7.96\% and MAGMA
by 28.00 percentage points.
It achieves the best performance on multi-session
and temporal reasoning questions, outperforming
TiMem by 8.12 and 6.77 percentage points,
respectively.
These gains further support the effectiveness
of combining specialized memory organization
with temporal indexing for complex retrieval.
ETAMem also demonstrates high construction
efficiency, requiring only 63.41k tokens
per session, 88.75\% and 67.70\% fewer than
TiMem and MAGMA, respectively.

\subsection{Ablation Studies}

\begin{table}[t]
\caption{Ablation study of key components.}
\label{tab:ablation}
\centering
\small
\setlength{\tabcolsep}{6pt}
\resizebox{0.7\columnwidth}{!}{%
\begin{tabular}{lcc}
\toprule
\textbf{Method}
& \textbf{LoCoMo}
& \textbf{LongMemEval-S} \\
\midrule
\textbf{ETAMem}
& \textbf{84.68}
& \textbf{83.00} \\
\midrule
w/o AMU
& 82.14
& \underline{80.60} \\
w/o Memory Tree
& 82.86
& \underline{80.60} \\
w/o Fallback
& \underline{84.48}
& 74.20 \\
\bottomrule
\end{tabular}%
}
\end{table}
\begin{table*}[t]
\caption{Effect of view-specific memory organization on LoCoMo and LongMemEval-S.}
\label{tab:organization}
\centering
\small
\resizebox{0.9\textwidth}{!}{
\begin{tabular}{lccccc|ccccccc}
\toprule
\multirow{2}{*}{\textbf{Method}}
& \multicolumn{5}{c|}{\textbf{LoCoMo}}
& \multicolumn{7}{c}{\textbf{LongMemEval-S}} \\
\cmidrule(lr){2-6}
\cmidrule(lr){7-13}
& \textbf{Overall}
& \textbf{Single}
& \textbf{Temp.}
& \textbf{Open}
& \textbf{Multi}
& \textbf{Overall}
& \textbf{KU}
& \textbf{MS}
& \textbf{SSA}
& \textbf{SSP}
& \textbf{SSU}
& \textbf{TR} \\
\midrule

\textbf{ETAMem}
& \textbf{84.68}
& \textbf{88.47}
& \textbf{80.06}
& \textbf{80.21}
& \textbf{80.14}
& \textbf{83.00}
& \textbf{87.18}
& \textbf{78.95}
& \textbf{92.86}
& \textbf{76.67}
& \textbf{95.71}
& \textbf{75.19} \\

\midrule

Unified Semantic
& \underline{83.31}
& \underline{87.75}
& 76.64
& \underline{77.08}
& \underline{79.79}
& 80.49
& \underline{84.61}
& \underline{76.69}
& 85.71
& 73.33
& \underline{95.71}
& 73.28 \\

Unified Affective
& 77.20
& 80.85
& 71.34
& 66.67
& 76.60
& \underline{80.60}
& 83.33
& 74.44
& \underline{92.86}
& \underline{76.67}
& 92.86
& 74.44 \\

Unified Temporal
& 82.92
& 86.92
& \underline{77.88}
& 75.00
& 79.43
& 79.20
& 80.76
& 73.68
& 82.14
& \underline{76.67}
& 94.29
& \underline{75.19} \\

\bottomrule
\end{tabular}
}
\end{table*}

\begin{table}[t]
\centering
\caption{Effect of view-specific memory representations on LongMemEval-S using Qwen3-30B.}
\label{tab:representation}
\small
\setlength{\tabcolsep}{4pt}
\renewcommand{\arraystretch}{1.08}
\resizebox{\columnwidth}{!}{
\begin{tabular}{lccccccc}
\toprule
\textbf{Method}
& \textbf{Overall}
& \textbf{KU}
& \textbf{MS}
& \textbf{SSA}
& \textbf{SSP}
& \textbf{SSU}
& \textbf{TR} \\
\midrule

Shared Text
& 72.80
& \underline{78.21}
& \underline{69.92}
& \underline{89.29}
& 73.33
& 92.85
& 54.89 \\

Shared Affect
& \underline{74.40}
& 76.92
& \underline{69.92}
& \underline{89.29}
& \underline{76.67}
& \underline{94.28}
& \underline{60.15} \\

\midrule

\textbf{ETAMem}
& \textbf{80.40}
& \textbf{87.18}
& \textbf{70.68}
& \textbf{91.07}
& \textbf{93.33}
& \textbf{95.71}
& \textbf{70.68} \\

\bottomrule
\end{tabular}
}
\end{table}

\begin{table}[t]
\caption{
Comparison of memory view configurations and
retrieval routing strategies using Qwen3-30B.
}
\label{tab:routing_ablation}
\centering
\small
\setlength{\tabcolsep}{8pt}
\begin{tabular}{lcc}
\toprule
\textbf{Configuration}
& \textbf{LoCoMo}
& \textbf{LongMemEval-S} \\
\midrule

\textbf{ETAMem}
& \textbf{80.84}
& \textbf{80.40} \\
\midrule

\multicolumn{3}{l}{\textit{Route Configuration}} \\

Event-only
& 73.44
& {74.00} \\

Time-only
& 77.73
& 74.40 \\

Affect-only
& 56.23
& 37.40 \\

Event + Time
& 77.92
& \underline{74.80} \\

Event + Affect
& 76.62
& 74.20 \\

Time + Affect
& \underline{79.35}
& 74.40 \\

\midrule
\multicolumn{3}{l}{\textit{Routing Strategy}} \\

Max
& 72.86
& 64.20 \\

Mean
& 79.22
& 75.00 \\
LLM-based
& 73.57
& 74.80 \\
\bottomrule
\end{tabular}

\end{table}

\subsubsection{Key Components}

Table~\ref{tab:ablation} presents the ablation
results of key components.
Specifically, \textit{w/o AMU} replaces atomic
memory units with dialogue turns,
\textit{w/o Memory Tree} removes hierarchical
memory organization, and
\textit{w/o Fallback} disables fallback
when retrieved evidence is insufficient.
Removing AMU extraction reduces accuracy from
84.68\% to 82.14\% on LoCoMo and from 83.00\%
to 80.60\% on LongMemEval-S, demonstrating
the benefit of fine-grained memory representation.
Removing hierarchical organization decreases
accuracy to 82.86\% and 80.60\%, respectively,
confirming the value of structured memory.
Disabling fallback causes a larger performance
drop on LongMemEval-S than on LoCoMo.
This is mainly because its lengthy dialogues
contain detailed narratives that may be
compressed during AMU extraction.
Fallback allows the model to recover missing
details from the original conversations
when retrieved memories are insufficient.

\subsubsection{Effect of View-Specific Memory Organization}

To investigate whether heterogeneous dialogue
information benefits from specialized organization,
we compare ETAMem with three unified strategies
that organize all AMUs into a single hierarchy
without distinguishing information types.
\textit{Unified Semantic} and
\textit{Unified Affective} organize memories
using textual and affective embeddings,
respectively, while
\textit{Unified Temporal} employs a unified
temporal hierarchy.
As shown in Table~\ref{tab:organization},
all unified strategies underperform ETAMem
in overall accuracy.
Unified Semantic performs best among the
unified strategies on LoCoMo, while
Unified Affective achieves the highest
overall accuracy on LongMemEval-S.
Notably, Unified Affective matches ETAMem
on preference-related questions (SSP),
while Unified Temporal matches ETAMem
on temporal reasoning (TR).
These results suggest that different
organizational principles can benefit
specific information types, but no single
unified structure effectively accommodates
their heterogeneous characteristics.
By integrating complementary type-specific
memory views, ETAMem achieves superior
overall performance.

\subsubsection{Effect of View-Specific Memory Representations}

We further investigate whether separating
information types is sufficient, or whether
different memory views also benefit from
specialized representations.
To this end, we compare ETAMem with two
variants that retain the Event and Affect
views but use shared representations.
\textit{Shared Text} uses textual embeddings
for both views, whereas
\textit{Shared Affect} uses affective
embeddings for both views.
The Time view 
remains unchanged.
As shown in Table~\ref{tab:representation},
both variants underperform ETAMem,
with overall accuracy decreasing by
7.60 and 6.00 percentage points,
respectively.
These results indicate that distinguishing
information types alone does not fully
capture their heterogeneous characteristics.
View-specific representations further
enhance the effectiveness of specialized
memory organization.

\subsubsection{Effect of Memory View Configurations}

To investigate whether combining complementary
memory views improves retrieval performance,
we evaluate different view configurations
using Qwen3-30B.
As shown in Table~\ref{tab:routing_ablation},
Time-only achieves the strongest performance
among single-view configurations on both
benchmarks.
This may be attributed to its complete
memory coverage, whereas the Event and Affect
views each contain only a subset of AMUs.
Combining views generally improves performance,
with Time + Affect achieving 79.35\%
on LoCoMo and Event + Time reaching
74.80\% on LongMemEval-S.
ETAMem further achieves 80.84\% and
80.40\% on the two benchmarks,
respectively, suggesting the benefits
of jointly retrieving from complementary
memory views.

\subsubsection{Effect of Retrieval Routing Strategies}

To evaluate the effectiveness of query-adaptive
budget allocation, we compare ETAMem with
three alternative routing strategies using
Qwen3-30B.
\textit{Max Routing} selects the highest-weighted
view, \textit{Mean Routing} assigns equal
retrieval budgets across views, and
\textit{LLM Routing} uses an LLM to select
the most appropriate view.
As shown in Table~\ref{tab:routing_ablation},
Mean Routing achieves the strongest performance
among the alternatives, with accuracies of
79.22\% and 75.00\% on LoCoMo and
LongMemEval-S, respectively.
ETAMem further improves accuracy to
80.84\% and 80.40\%, outperforming
Mean Routing by 1.62 and 5.40
percentage points, respectively.
These results support the effectiveness
of adaptively allocating retrieval budgets
across complementary memory views.

\subsection{Further Analysis}
We further investigate ETAMem through affect
analysis, hyperparameter sensitivity,
and a case study.
Additional analyses in
Appendix~\ref{app:experimental_analysis}
include human evaluation of affect classification,
memory visualizations,
routing weight distributions, and error analysis.

\subsubsection{Affect Analysis}
Since LoCoMo and LongMemEval-S do not provide
explicit annotations for affect-related question
categories, we further evaluate ETAMem on
CloneMem~\cite{hu-etal-2026-clonemem},
which includes three categories of personal
memory questions: Experience, Emotion, and Opinion.
Detailed settings are provided
in Appendix~\ref{sec:dataset_details}.

\begin{table}[t]
\caption{
Performance comparison on CloneMem.
Choice accuracy is reported on a 0--100 scale,
while memory helpful score and memory recall
are reported on a 0--1 scale.
}
\label{tab:clonemem}
\centering
\small
\resizebox{0.9\columnwidth}{!}{
\begin{tabular}{lccc}
\toprule
\textbf{Method}
& \textbf{Choice Acc.}
& \textbf{Memory Helpful}
& \textbf{Memory Recall} \\
\midrule

A-MEM
& 87.48 & 0.6275 & 0.5490 \\

Mem0
& 85.28 & 0.6182 & 0.4860 \\

Flat
& 87.74 & 0.7767 & 0.6492 \\

\midrule

\textbf{ETAMem}
& \textbf{92.59}
& \textbf{0.9200}
& \textbf{0.8031} \\

\bottomrule
\end{tabular}
}
\end{table}

As shown in Table~\ref{tab:clonemem},
ETAMem outperforms all baselines across
three metrics, achieving 92.59\% choice accuracy,
0.9200 memory helpfulness, and 0.8031 recall,
with improvements of 4.85 percentage points,
18.45\%, and 23.71\%, respectively.
These results demonstrate its effectiveness
in retrieving relevant and useful memories.

\begin{figure}[t]
\centering
\includegraphics[width=0.92\columnwidth]
{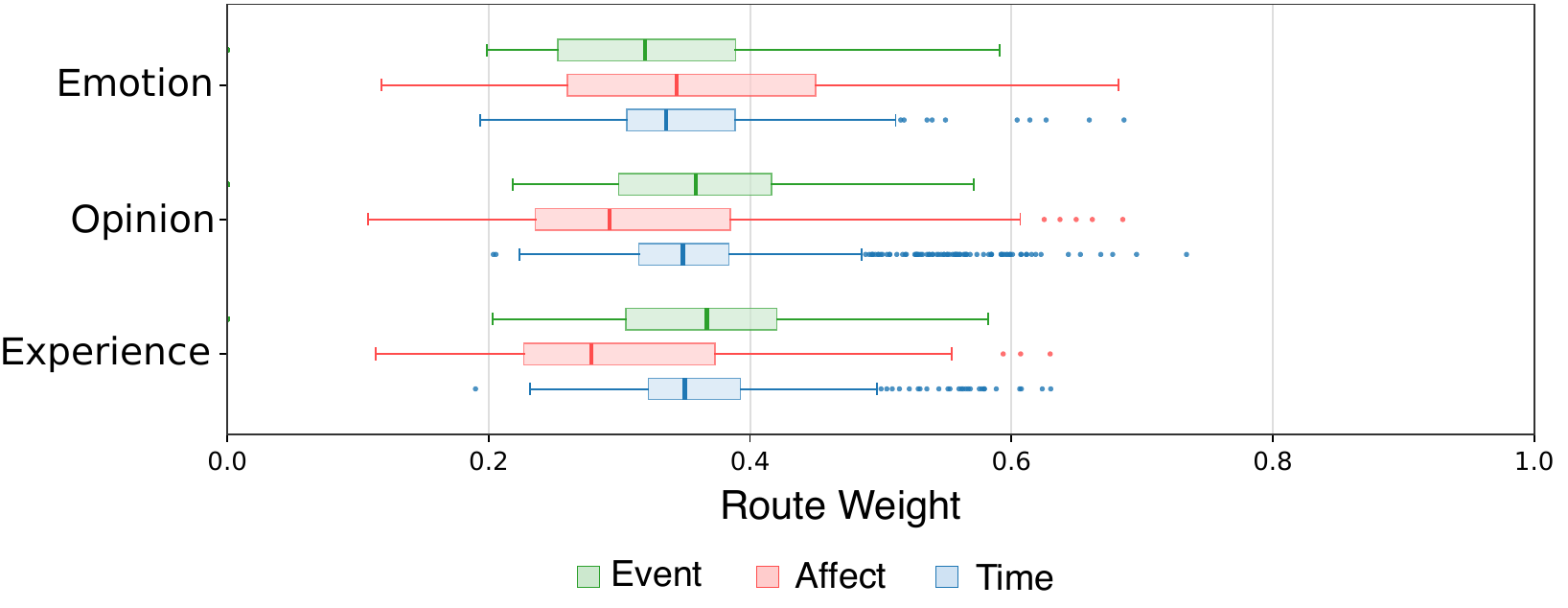}
\caption{
Distribution of routing weights across
Experience, Emotion, and Opinion questions
in CloneMem.
}
\label{fig:clonemem_routing}
\end{figure}

To examine the contributions of different
memory views, we analyze routing weights
across three question categories.
As shown in Figure~\ref{fig:clonemem_routing},
the Affect view receives a slightly higher
median weight for Emotion questions,
indicating its relevance to emotional information.
Experience questions favor the Event view,
reflecting their reliance on episodic memories.
For Opinion questions, Event and Time
receive higher weights, highlighting
the importance of factual and temporal context.
These results demonstrate the complementary
roles of different memory views.

\begin{figure}[t]
    \centering
    \includegraphics[width=0.9\linewidth]{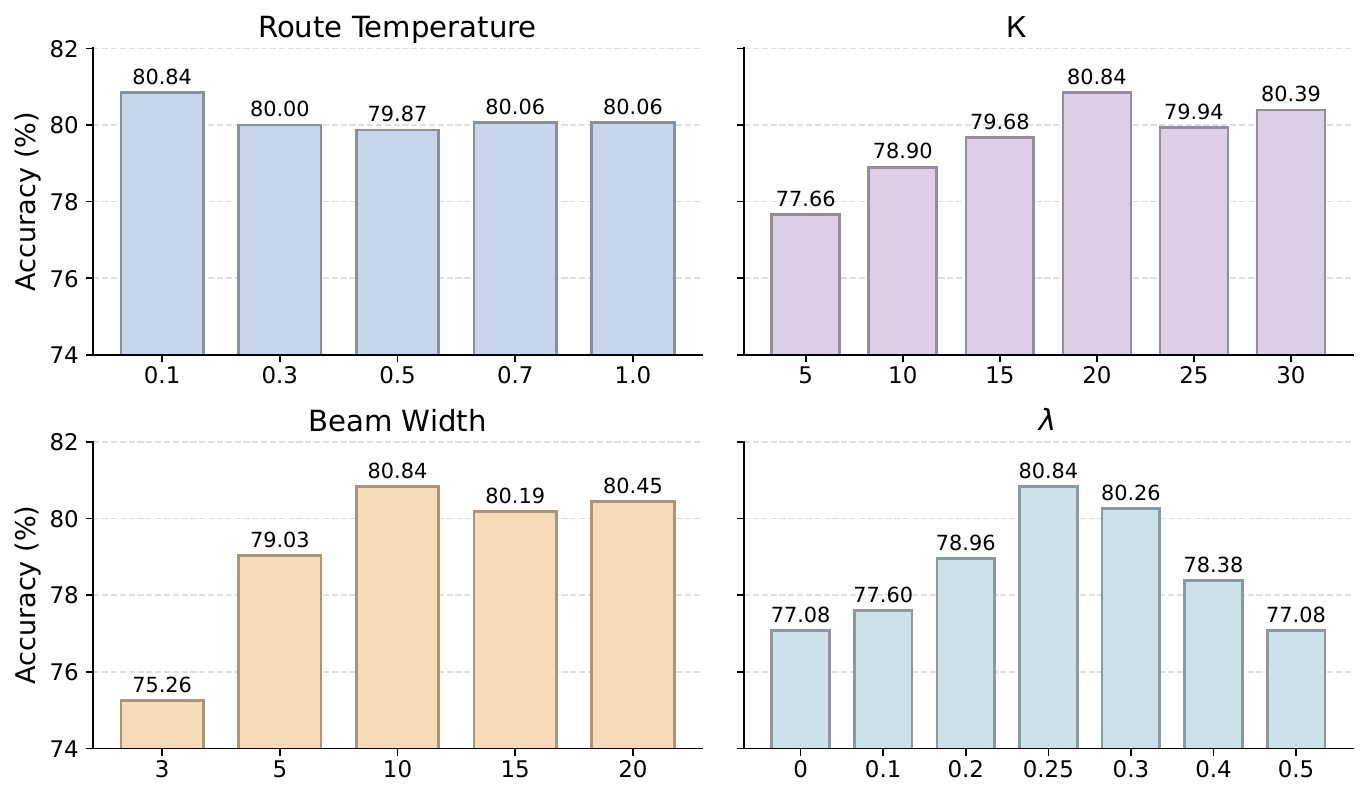}
\caption{Hyperparameter sensitivity on LoCoMo.}
    \label{fig:hyperparameter}
\end{figure}
\subsubsection{Hyperparameter Sensitivity}

We analyze the sensitivity of ETAMem to four hyperparameters: routing temperature
$\beta$, retrieval budget $K$, beam width $B$,
and lexical coefficient $\lambda$.
All experiments in this section are conducted
with Qwen3-30B as the backbone.

As shown in Figure~\ref{fig:hyperparameter},
the best performance is achieved with
$\beta=0.1$, $K=20$, and $B=10$,
indicating that a concentrated routing
distribution, sufficient retrieval coverage,
and moderate hierarchical exploration
are beneficial.
For $\lambda$, incorporating lexical matching
consistently improves performance.

% \begin{table}[t]
% \centering
% \scriptsize
% \setlength{\tabcolsep}{3pt}
% \renewcommand{\arraystretch}{1.10}

% \begin{tabularx}{\columnwidth}{
%     @{}
%     >{\bfseries\raggedright\arraybackslash}p{0.19\columnwidth}
%     >{\raggedright\arraybackslash}X
%     @{}
% }
% \toprule

% Question &
% \emph{How many days passed between the day I started playing
% along to my favorite songs on my old keyboard and the day I
% discovered a bluegrass band?}
% \\

% \midrule

% View Routing &
% Event: weight 0.263;\quad
% Affect: weight 0.466;\quad
% Time: weight 0.271
% \\

% \midrule

% Event Tree &
% musical interests and instruments
% $\rightarrow$
% piano-featured music and exploration of bluegrass
% $\rightarrow$
% playing favorite songs on an old keyboard
% $+$
% discussing and exploring bluegrass music
% \\

% \midrule

% Affect Tree &
% enjoyable activities
% $\rightarrow$
% enjoyable keyboard playing and a newfound appreciation for
% bluegrass
% $\rightarrow$
% playing along was fun and motivated further practice
% $+$
% the user started enjoying bluegrass music
% \\

% \midrule

% Time Tree &
% \texttt{WEEK\_2023-W12}
% $\rightarrow$
% \texttt{DAY\_2023-03-25}
% $\rightarrow$ 
% keyboard-playing session
% \\

% &
% \texttt{WEEK\_2023-W13}
% $\rightarrow$
% \texttt{DAY\_2023-03-31}
% $\rightarrow$
% bluegrass-discovery session
% \\

% \midrule

% Answer &
% March 31, 2023 $-$ March 25, 2023
% $= \mathbf{6\ days}$
% \\
% \bottomrule
% \end{tabularx}
% \caption{A multi-hop case from LoCoMo.}
% \label{tab:case}
% \end{table}

\begin{table}[t]
\caption{A case study of a knowledge-updating question in LongMemEval-S.}
\label{tab:case}
\centering
\scriptsize
\setlength{\tabcolsep}{3pt}
\renewcommand{\arraystretch}{1.10}

\begin{tabularx}{\columnwidth}{
    @{}
    >{\bfseries\raggedright\arraybackslash}p{0.19\columnwidth}
    >{\raggedright\arraybackslash}X
    @{}
}
\toprule

Question &
\emph{What brand of BBQ sauce am I currently obsessed with?}
\\

\midrule

View Routing &
Event: weight 0.284;\quad
Affect: weight 0.452;\quad
Time: weight 0.264
\\

\midrule

Event Tree &
BBQ cooking activities
$\rightarrow$
brisket, ribs, pulled pork, and sauce recommendations
$\rightarrow$
general BBQ sauce alternatives, including
Sweet Baby Ray's BBQ sauce
\\

\midrule

Affect Tree &
food preferences and enthusiasm
$\rightarrow$
favorite and currently preferred BBQ sauces
$\rightarrow$
the user is currently obsessed with
Kansas City Masterpiece BBQ sauce
\\

\midrule

Time Tree &
\texttt{WEEK\_2023-26}
$\rightarrow$
\texttt{DAY\_2023-06-30}
$\rightarrow$
BBQ cooking and sauce-preference session
\\

\midrule

Answer &
The user is obsessed with
\textbf{Kansas City Masterpiece BBQ sauce}.
\\

\bottomrule
\end{tabularx}
\end{table}

% \begin{table}[t]
% \centering
% \scriptsize
% \setlength{\tabcolsep}{3pt}
% \renewcommand{\arraystretch}{1.10}

% \begin{tabularx}{\columnwidth}{
%     @{}
%     >{\bfseries\raggedright\arraybackslash}p{0.19\columnwidth}
%     >{\raggedright\arraybackslash}X
%     @{}
% }
% \toprule

% Question &
% \emph{What type of rice is my favorite?}
% \\

% \midrule

% View Routing &
% Event: weight 0.267;\quad
% Affect: weight 0.450;\quad
% Time: weight 0.283
% \\

% \midrule

% Event Tree &
% personal interests and activities
% $\rightarrow$
% food-related discussions and Japanese-style dishes
% $\rightarrow$
% the user was thinking of making some Japanese-style dishes with Japanese short-grain rice
% \\

% \midrule

% Affect Tree &
% personal preferences and interests
% $\rightarrow$
% food and lifestyle preferences
% $\rightarrow$
% interest in Japanese short-grain rice
% \\

% \midrule

% Time Tree &
% \texttt{WEEK\_2023-W21}
% $\rightarrow$
% \texttt{DAY\_2023-05-27}
% $\rightarrow$
% Japanese-style cooking and onigiri-preparation session
% \\

% \midrule

% Answer &
% The user's favorite type of rice is
% \textbf{Japanese short-grain rice}.
% \\

% \bottomrule
% \end{tabularx}
% \caption{A preference-oriented case from LongMemEval.}
% \label{tab:case}
% \end{table}

\subsubsection{Case Study}
\label{sec:case-study}

To analyze how different memory views contribute to retrieval, we present a representative knowledge-update case from LongMemEval in Table~\ref{tab:case}.
For this query, ETAMem assigns the highest routing weight to the Affect view, as the answer depends on the user's preference rather than merely factual events.
The Event view retrieves related BBQ cooking experiences and sauce-related information, while the Affect view captures the user's current preference for a specific BBQ sauce.
The Time view further helps locate the relevant memory update over time.
By integrating complementary evidence from different views, ETAMem accurately identifies the user's current preference.

\section{Conclusion}

In this paper, we proposed ETAMem, a multi-view
memory framework that decouples heterogeneous
conversational information and aligns memory
representations and organization with its
distinct characteristics.
ETAMem integrates heterogeneous information
parsing, view-specific memory structuring,
and entropy-guided  retrieval,
supported by dynamic maintenance and adaptive
calibration.
Experiments on two long-term memory benchmarks
demonstrate relative improvements of 12.46\% and
7.96\% over the strongest baselines, respectively,
with strong multi-hop performance and competitive
token efficiency.
Further analysis highlights the complementary
benefits of different memory views and the
effectiveness of adaptive routing.
Despite these improvements, our affective representations rely on sentiment polarity, which may fail to identify neutral expressions of preferences or attitudes as affective information. Future work will incorporate explicit affective relevance classification to address this limitation.

%
% Despite these improvements, ETAMem relies on
% off-the-shelf affective models, which could benefit
% from dialogue-specific fine-tuning.
% Moreover, ETAMem does not explicitly model memory
% forgetting.
% Future work will explore adaptive memory evolution
% through forgetting and pruning strategies.

\bibliographystyle{ACM-Reference-Format}
\bibliography{sample-base}

\appendix
% \section{Method Details}
% \label{method_details}
% \subsection{Notation}
% \label{app:parameter_settings}
% \input{Figures/notation}

% \subsection{Algorithms}
% \label{app:algorithms}

% Algorithms~\ref{alg:etamem-construction},
% \ref{alg:etamem-maintenance}, and
% \ref{alg:etamem-retrieval} provide an  overview of ETAMem.
% %
% Algorithm~\ref{alg:etamem-construction} describes online memory
% construction, including AMU extraction, representation, and insertion.
% %
% Algorithm~\ref{alg:etamem-maintenance} presents session-level
% maintenance of the Event and Affect views through splitting, merging, and lazy summary
% update.
% %
% Algorithm~\ref{alg:etamem-retrieval} details retrieval and
% inference, including entropy-based budget allocation, layer-wise beam
% search, and fallback to the source
% dialogue turns associated with the retrieved AMUs when the
% evidence is insufficient.

% \newpage
% \newpage
% \input{Figures/construction_alg}
% \input{Figures/inference_alg}

\section{Complexity Analysis}
Let $A=\sum_{t=1}^{T}n_t$ denote the number of AMUs,
$H$ the maximum tree depth,
$L=L_{\mathrm{event}}+L_{\mathrm{affect}}+L_{\mathrm{time}}$ the number of leaf nodes,
$b_{\max}$ the maximum number of children of any internal node,
$R$ the number of root nodes per view,
$N$ the number of nodes (leaf and internal) in the Event and Affect views,
and $D$ the embedding dimension.
We count LLM calls and dominant similarity and clustering operations, assuming cached centroids and summary embeddings.

\textbf{LLM Calls.}
AMU extraction requires one LLM call per turn, whereas leaf summaries are built directly from AMU texts without LLM calls.
Each insertion, split, or merge increments the pending-change counters of the affected nodes and at most $H$ ancestors, and a summary is regenerated once its counter reaches $\delta$.
Let $J$ count initial internal-node summary generations and $U$ the total number of counter increments, where insertions contribute $O(AH)$.
Thus, memory construction requires $O(T+J+U/\delta)$ LLM calls.

\textbf{Memory Management.}
For the Event and Affect views, beam-search insertion costs $O\big(Db_{\max}(R+BH)\big)$ per AMU, i.e., $O\big(ADb_{\max}(R+BH)\big)$ in total.
Time-view insertion follows temporal indexing with $O(A)$ total cost.
At the end of each session, ETAMem checks for potential Split and Merge operations in the Event and Affect views.
For a leaf with $m$ AMUs, the cohesion check costs $O(m^{2}D)$, and if a split is triggered, cosine $K$-means adds $O(IkmD)$ for $k$ clusters and $I$ iterations; here $m$ is the leaf size before splitting and may exceed $M_{\max}$.
For Merge, computing pairwise centroid similarities among the $N$ nodes costs $O(N^{2}D)$ per maintenance step in the worst case.

\textbf{Retrieval.}
View routing computes similarities over all leaf summaries with $O(LD)$ cost.
Beam search costs $O\big(Db_{\max}(R+BH)\big)$ per view.
Since the number of views is a constant (three), this factor is omitted, and the retrieval complexity is $O\big(D[L+b_{\max}(R+BH)]\big)$, excluding BM25 scoring, encoder inference, and LLM computation.

% \section{Complexity Analysis}
% Let $A=\sum_{t=1}^{T}n_t$ denote the number of AMUs,
% $H$ the maximum tree depth,
% $L=L_{\mathrm{event}}+L_{\mathrm{affect}}+L_{\mathrm{time}}$ the number of leaf nodes,
% $b_{\max}$ the maximum number of children of any internal node,
% and $N$ the number of internal nodes.

% \textbf{LLM Calls.}
% AMU extraction requires one LLM call per dialogue turn.
% Each update affects at most $H$ ancestor nodes, and summary regeneration is triggered every $\delta$ updates.
% Thus, memory construction requires $O(T+AH/\delta)$ LLM calls.

% \textbf{Memory Management.}
% For the Event and Affect views, layer-wise beam search expands at most $B$ nodes and examines at most $b_{\max}$ children per node at each level, resulting in $O(Bb_{\max}H)$ cost per AMU and $O(ABb_{\max}H)$ in total.
% Time-view insertion follows temporal indexing with $O(A)$ total cost.
% %
% At the end of each session, structural maintenance performs Split and Merge.
% Split costs $O(AM_{\max})$, since cohesion computation is performed within each leaf and each leaf contains at most $M_{\max}$ AMUs.
% Merge operates on internal nodes and computes pairwise similarities among $N$ internal node centroids, resulting in $O(N^{2}D_{\mathrm{emb}})$ time complexity, where $D_{\mathrm{emb}}$ denotes the embedding dimension.

% \textbf{Retrieval.}
% View routing computes similarities over all leaf summaries with $O(L)$ cost.
% Beam search costs $O(Bb_{\max}H)$ per view.
% Thus, the retrieval complexity is
% $O(L+Bb_{\max}H)$.

\section{Experimental Setup}
\label{sec:dataset_details}
\subsection{Dataset Details}
\textbf{LoCoMo}.
LoCoMo~\cite{locomo} evaluates long-term conversational
memory through multi-session dialogues with temporal information. We use
the full benchmark containing 10 conversation groups and 1,540 questions,
covering single-hop retrieval, multi-hop reasoning, open-domain queries,
and temporal reasoning.

\textbf{LongMemEval-S.}
LongMemEval-S~\cite{longmemeval} evaluates memory systems under
long-term interactions. It contains 500 conversations and 500 questions,
covering knowledge updates, multi-session reasoning, single-session
interactions, temporal reasoning, and selective response generation.

\textbf{CloneMem.}
CloneMem~\cite{hu-etal-2026-clonemem} is
a long-term personal memory benchmark
constructed from non-conversational digital
traces, including diaries, social media posts,
and emails spanning one to three years.
CloneMem evaluates memory capabilities across
three dimensions: Experience, Emotion, and
Opinion.

\subsection{Evaluation Protocol}
\textbf{Implementation Details.}
Following TiMem~\cite{timem}, we adopt the same
experimental settings for fair comparison.
Baseline results in Tables~\ref{locomo} and
\ref{longmemeval_results} are taken from
TiMem~\cite{timem}, except for LightMem and MAGMA,
which we evaluate under the same settings.
Following the CloneMemBench evaluation protocol,
we set the retrieval budget to $K=5$ and use
GPT-4o-mini as the backbone LLM and
text-embedding-3-small as the embedding model.
We compare ETAMem against A-MEM~\cite{amem}, Mem0~\cite{mem0}, and Flat Retrieval,
with results taken from
CloneMem~\cite{hu-etal-2026-clonemem}.

\textbf{Evaluation Metrics.}
We adopt three evaluation metrics from CloneMem:
Choice Accuracy, Memory Helpfulness, and
Memory Recall.
Choice Accuracy measures the percentage of
correctly answered multiple-choice questions.
Memory Helpfulness evaluates the usefulness
of retrieved evidence for answering questions,
while Memory Recall measures the semantic
coverage of ground-truth memory information.
The latter two metrics are reported on a
normalized scale from 0 to 1.

\begin{table*}[t]
\caption{Representative partial and complete retrieval failures.}
\label{tab:error_cases}
\centering
\scriptsize
\setlength{\tabcolsep}{3pt}
\renewcommand{\arraystretch}{1.0}

\begin{tabular}{
    @{}
    >{\bfseries\raggedright\arraybackslash}p{0.13\textwidth}
    >{\raggedright\arraybackslash}p{0.26\textwidth}
    >{\raggedright\arraybackslash}p{0.18\textwidth}
    >{\raggedright\arraybackslash}p{0.36\textwidth}
    @{}
}
\toprule
Failure Mode &
Question and Outcome &
View Routing &
Retrieved Path and Evidence Coverage
\\
\midrule

Partial Retrieval Failure
&
\textbf{Dataset:} LongMemEval

\textbf{Question:}
\emph{How many different types of citrus fruits have I used in my
cocktail recipes?}

\textbf{Gold:}
3

\textbf{Prediction:}
2 (orange and lemon)
&
Event: 0.333

Affect: 0.147

Time: 0.519
&
\textbf{Event view:}
culinary experiences
$\rightarrow$
cocktail recipes

\textbf{Time view:}
\texttt{WEEK\_2023-W21}
$\rightarrow$
\texttt{DAY\_2023-05-21}
$\rightarrow$
orange-bitters session

\textbf{Coverage:}
the orange- and lemon-related evidence is retained, but the May 25
Daiquiri memory containing fresh lime juice is omitted; recall $=0.75$.
\\

\midrule

Retrieval Miss
&
\textbf{Dataset:} LoCoMo

\textbf{Question:}
\emph{What book did Caroline recommend to Melanie?}

\textbf{Gold:}
\emph{Becoming Nicole}

\textbf{Prediction:}
The specific book is not mentioned
&
Event: 0.613

Affect: 0.147

Time: 0.239
&
\textbf{Event view:}
personal growth and art
$\rightarrow$
pottery and creative activities

\textbf{Affect view:}
support and encouragement
$\rightarrow$
self-acceptance

\textbf{Time view:}
July 2023 sessions

\textbf{Coverage:}
the recommendation turn containing \emph{Becoming Nicole} is absent
after both dialogue fallback; recall $=0$.
\\

\bottomrule
\end{tabular}

\end{table*}

\section{Additional Analysis}
\label{app:experimental_analysis}

\subsection{Human Evaluation of Affect Classification}
\label{app:affect_evaluation}

To assess affect identification reliability,
we sampled 400 Atomic Memory Units (AMUs)
from LoCoMo and LongMemEval-S.
Two university annotators voluntarily and
anonymously labeled each AMU as affective
or non-affective, independently following
shared guidelines.
Cohen's kappa was 0.651, and disagreements
were resolved through discussion to obtain
gold-standard labels.
The guidelines are provided in
Table~\ref{tab:human_affect_guidelines}.

\begin{table}[t]
\caption{Annotation guidelines for
affect classification.}
\label{tab:human_affect_guidelines}
\centering
\small
\setlength{\tabcolsep}{8pt}
\renewcommand{\arraystretch}{1.15}

\begin{tabularx}{\columnwidth}{@{}X@{}}
\toprule

Given a memory unit, determine whether
it contains affective information based solely on
its explicitly expressed meaning. \\[6pt]

\textbf{Affective (1):}
The memory unit expresses emotions, psychological feelings,
personal preferences, likes or dislikes, or
sentiment-bearing subjective evaluations and attitudes. \\[6pt]

\textbf{Non-Affective (0):}
The memory unit describes neutral events, factual information,
objective attributes, general plans, task-oriented
intentions, or information requests without explicit
affective content. \\[6pt]

\textbf{Guidelines:}
Judge only the information expressed in the memory unit.
Do not infer emotions from surrounding dialogue
or the potential implications of an event.
Distinguish emotional expectations from general
plans, and subjective evaluations from objective
descriptions. \\[6pt]

\textbf{Output:}
Assign exactly one binary label (0 or 1) to each memory unit. \\

\bottomrule
\end{tabularx}

\end{table}

\textbf{LLM-based Affect Classification.}
For comparison, we employ GPT-4o-mini to identify
affective information using the prompt shown in
Table~\ref{tab:affective_classification_prompt}.
The model produces a binary response
(\textit{yes} or \textit{no}), corresponding
to affective and non-affective labels, respectively.

\begin{table}[t]
\caption{Prompt for GPT-4o-mini-based
affect classification.}
\label{tab:affective_classification_prompt}
\centering
\small
\setlength{\tabcolsep}{4pt}
\renewcommand{\arraystretch}{1.05}

\begin{tabularx}{\columnwidth}{@{}X@{}}
\toprule

Determine whether the atomic fact expresses a person's
subjective affect, such as emotion, attitude, preference,
like/dislike, desire, or evaluation. \\

Answer "yes" if it contains subjective affective
information. \newline
Answer "no" if it is only an objective or factual
statement. \\

Do not infer emotions that are not explicitly stated. \\

\textbf{Return only:}
yes or no \\

\bottomrule
\end{tabularx}

\end{table}

\textbf{Classification Performance.}
Table~\ref{tab:affect_classification}
compares RoBERTa-Sentiment with GPT-4o-mini
on the human-annotated benchmark.
RoBERTa-Sentiment achieves an F1 score of
80.53\%, surpassing GPT-4o-mini at 79.39\%.
It also achieves higher precision of 80.79\%,
compared with 73.60\% for GPT-4o-mini,
although its recall is lower at 80.26\%
versus 86.18\%.
These results demonstrate that a lightweight
classifier provides competitive affect
identification without additional LLM
inference during memory construction.

\begin{table}[t]
\caption{Affect classification performance.}
\label{tab:affect_classification}
\centering
\small
\setlength{\tabcolsep}{8pt}
\renewcommand{\arraystretch}{1.08}

\begin{tabular}{lccc}
\toprule
Method & Precision & Recall & F1 \\
\midrule
GPT-4o-mini
& 73.60 & \textbf{86.18} & 79.39 \\
RoBERTa-Sentiment
& \textbf{80.79} & 80.26 & \textbf{80.53} \\
\bottomrule
\end{tabular}

\end{table}

\subsection{Memory Visualization}
\label{app:tree_visualization}

\subsubsection{Memory Structure Visualization}
\begin{figure}[t]
    \centering
    \includegraphics[width=0.9\linewidth]
    {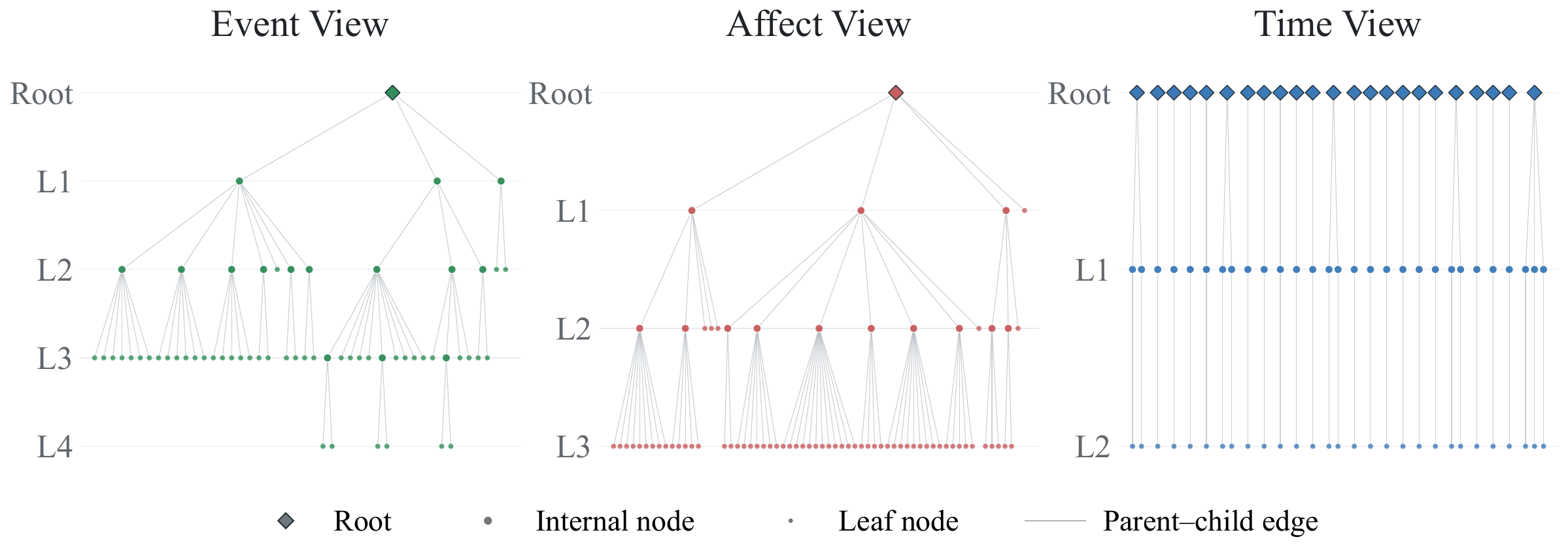}
    \caption{Visualization of the memory structure on
    LoCoMo.}
    \label{fig:locomo_tree}
\end{figure}

To illustrate the structural differences among the three memory views, we
visualize Event, Affect, and Time views from LoCoMo.
As shown in
Figure~\ref{fig:locomo_tree}, the three
views organize the conversational memory with distinct hierarchical
structures.
The Event and Affect views form tree-structured semantic organizations with
different branching patterns, while the Time view exhibits a shallow forest
structure with multiple roots.

\begin{figure*}[t]
    \centering
    \includegraphics[width=0.9\linewidth]{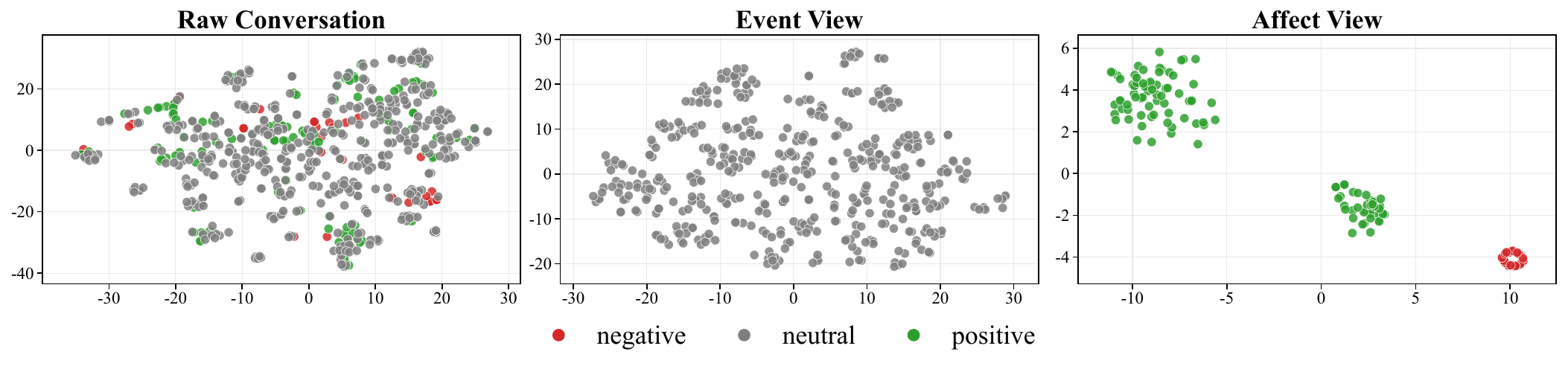}
\caption{T-SNE visualization of LongMemEval memory representations under raw conversation, event, and affect view.}
    \label{fig:embedding_visualization}
\end{figure*}

% \begin{figure}[t]
%     \centering
%     \includegraphics[width=\linewidth]
%     {Figures/longmemeval_tree_visualization.pdf}
%     \caption{Visualization of the multi-view memory structure on
%     LongMemEval. }
%     \label{fig:longmemeval_tree}
% \end{figure}

\subsubsection{Representation Visualization}

We visualize representative memory representations from LongMemEval using t-SNE~\cite{van2008visualizing}. 
We exclude the temporal view, as it is organized by temporal hierarchies rather than representation similarity.
As shown in Figure~\ref{fig:embedding_visualization}, the affect view exhibits clear clustering patterns corresponding to different affective polarities, while the event view shows more concentrated distributions among event-related memories. These patterns suggest that ETAMem decouples heterogeneous memory signals that are originally mixed in dialogue representations, leading to more coherent clustering within each memory view.

% \subsection{Retrieval Routing Analysis}
% \label{sec:routing_analysis}

% \begin{figure}[t]
%     \centering
%     \includegraphics[width=0.45\textwidth]
%     {Figures/locomo_route.pdf}
%     \caption{Distribution of routing weights on LoCoMo.}
%     \label{fig:routing_weight_locomo}
% \end{figure}

% \begin{figure}[t]
%     \centering
%     \includegraphics[width=0.45\textwidth]
%     {Figures/longmemeval_route.pdf}
%     \caption{Distribution of routing weights on LongMemEval.}
%     \label{fig:routing_weight_longmemeval}
% \end{figure}
% We analyze how ETAMem allocates retrieval
% budgets across different memory views.
% Figure~\ref{fig:routing_weight_longmemeval}
% shows the routing weight distributions
% across question categories on LongMemEval-S.
% %
% Multi-session and knowledge-update questions
% favor the Event view, reflecting their reliance
% on factual and episodic memories.
% In contrast, single-session-assistant questions
% assign greater weight to the Time view,
% highlighting the importance of session context.
% Other categories exhibit more balanced
% distributions, demonstrating ETAMem's ability
% to adapt retrieval budgets to different
% question types.

\subsection{Error Analysis}
\label{sec:error_analysis}

\begin{table}[H]
\caption{Error distribution (\%) by evidence retrieval
status.}
\label{tab:error_distribution}
\centering
\small
\setlength{\tabcolsep}{7pt}
\renewcommand{\arraystretch}{1.14}
\resizebox{\columnwidth}{!}{%
\begin{tabular}{
    @{}
    l
    l
    r
    r
    r
    r
    r
    r
    @{}
}
\toprule
\multirow{2}{*}{Dataset} &
\multirow{2}{*}{Question Type} &
\multicolumn{2}{c}{Retrieved} &
\multicolumn{2}{c}{Unretrieved} &
\multicolumn{2}{c}{Partially Retrieved}
\\
\cmidrule(lr){3-4}
\cmidrule(lr){5-6}
\cmidrule(lr){7-8}
& & Correct & Incorrect & Correct & Incorrect
& Correct & Incorrect
\\
\midrule

\multirow{7}{*}{LongMemEval}
& KU  & 87.2 & 12.8 & 0.0 & 0.0 & 0.0 & 0.0 \\
& MS  & 78.9 & 17.3 & 0.0 & 0.0 & 0.0 & 3.8 \\
& SSA & 91.1 & 7.1 & 1.8 & 0.0 & 0.0 & 0.0 \\
& SSP & 76.7 & 20.0 & 0.0 & 3.3 & 0.0 & 0.0 \\
& SSU & 95.7 & 4.3 & 0.0 & 0.0 & 0.0 & 0.0 \\
& TR  & 69.2 & 21.1 & 0.0 & 0.0 & 6.0 & 3.8 \\
\cmidrule(lr){2-8}
& \textbf{Overall}
& \textbf{81.2} & \textbf{14.8}
& \textbf{0.2} & \textbf{0.2}
& \textbf{1.6} & \textbf{2.0} \\

\midrule

\multirow{5}{*}{LoCoMo}
& Single Hop
& 79.17 & 4.83 & 7.8 & 6.3 & 1.5 & 0.4 \\
& Temporal
& 73.26 & 10.24 & 3.1 & 7.5 & 3.7 & 2.2 \\
& Open Domain
& 34.41 & 7.29 & 28.1 & 9.4 & 17.7 & 3.1 \\
& Multi Hop
& 29.04 & 4.66 & 4.6 & 7.8 & 46.5 & 7.4 \\
\cmidrule(lr){2-8}
& \textbf{Overall}
& \textbf{65.97} & \textbf{6.08}
& \textbf{7.50} & \textbf{7.02}
& \textbf{11.21} & \textbf{2.23} \\

\bottomrule
\end{tabular}
}
\end{table}

\subsubsection{Representative Error Cases}
\label{sec:representative_error_cases}
Table~\ref{tab:error_cases} illustrates two representative
retrieval failures.
\textbf{Partial Retrieval.}
In this case, ETAMem retrieves most
relevant evidence but misses the May 25 dialogue
containing the third fruit, leading to an incorrect
count.
\textbf{Retrieval Miss.}
In this case, the retrieved
memories contain related discussions but omit the
target recommendation, \emph{Becoming Nicole}.
Fallback retrieval also fails to recover the missing
evidence.

\subsubsection{Error Distribution}
\label{sec:error_distribution}

We further categorize predictions by evidence
retrieval status, as shown in
Table~\ref{tab:error_distribution}.
On LongMemEval-S, 14.8\% of all questions
involve incorrect answers despite successful
evidence retrieval, suggesting limitations
in evidence aggregation and reasoning.
In contrast, 60.4\% of LoCoMo errors involve
missing or incomplete evidence, indicating
that retrieval coverage remains a major challenge.

\section{Prompt Templates}
\label{sec:prompt_design}
\begin{promptbox}{AMU Extraction (LoCoMo)}
Extract atomic facts from the utterance.
The speaker is \promptvar{<speaker>}.

\textbf{Rules:}

- One fact per line, separated by \texttt{\textbackslash n}. \newline
- Attribute opinions and general statements to the speaker
(e.g. "believes", "thinks"). \newline
- Every fact MUST begin with a subject. \newline
- Third person only; replace all pronouns with actual names. \newline
- Include the cause or context of states when inferable. \newline
- Ignore chit-chat and filler.

\textbf{Input:}
\promptvar{<utterance>}

\textbf{Output:}
\end{promptbox}

\begin{promptbox}{AMU Extraction (LongMemEval-S)}
If the utterance is primarily a task, puzzle,
riddle, or Q\&A about a hypothetical problem
(including its solution), output ONE summary line:
"\promptvar{<speaker>} discussed [topic]."

Otherwise extract personal atomic facts about
\promptvar{<speaker>}:

- One fact per line. \newline
- Third person; replace all pronouns with actual names. \newline
- Every fact must begin with a subject. \newline
- Attribute opinions/beliefs to the speaker
(e.g. "\promptvar{<speaker>} thinks/believes ..."). \newline
- Include cause or context when inferable. \newline
- Skip chit-chat and filler.

\textbf{Input:}
\promptvar{<utterance>}

\textbf{Output:}
\end{promptbox}

\begin{promptbox}{Memory Summarization}
Summarize the following contents.
Requirements: no more than 256 words,
directly provide the summary.

1. \promptvar{<content_1>} \newline
2. \promptvar{<content_2>} \newline
... \newline
N. \promptvar{<content_N>}

Summary:
\end{promptbox}

\begin{promptbox}{Answer Generation (LoCoMo)}
You are an intelligent memory assistant tasked with
retrieving accurate information from conversation memories.

\textbf{\# CONTEXT}

You have access to retrieved atomic facts from the
conversation. These facts may be relevant to answering
the question.

\textbf{\# INSTRUCTIONS}

1. Carefully analyze all provided atomic facts. \newline
2. Pay special attention to the timestamps to determine
the answer. \newline
3. If the question asks about a specific event or fact,
look for direct evidence in the atomic facts. \newline
4. If the atomic facts contain contradictory information,
prioritize the most recent one. \newline
5. If the question involves relative time references,
such as 'last year', 'two months ago', or 'next month',
calculate the actual date based on the memory timestamp.
For example, if a memory from 4 May 2022 mentions
'went to India last year', then the trip occurred in 2021. \newline
6. Always convert relative time references to specific
dates, months, or years. For example, convert 'last year'
to '2022' or 'two months ago' to 'March 2023' based on
the memory timestamp. Do not answer temporal questions
with timestamp formats such as '2023-05-08'.
Answer naturally, such as '7 May 2023' or
'the week before 6 July 2023'. \newline
7. Focus only on the content of the memories from both
speakers. Do not confuse character names mentioned
in memories with the actual users who created those
memories. \newline
8. The answer should be less than 5-8 words.

\textbf{\# Relevant Atomic Facts:}
\promptvar{<retrieved atomic facts>}

\textbf{Question:} \promptvar{<question>}

\textbf{Answer:}
\end{promptbox}

\begin{promptbox}{Answer Generation (LongMemEval-S)}
Please answer the question based on the relevant
information.

Relevant Atomic Facts:
\promptvar{<retrieved atomic facts>}

If the answer cannot be inferred from the provided
context, output: no information.

Current Date: \promptvar{<question date>}

Question: \promptvar{<question>}
Answer:
\end{promptbox}

\begin{promptbox}{Answer Evaluation}
Your task is to label an answer to a question as
'CORRECT' or 'WRONG'.

You will be given:

1. A question \newline
2. A gold answer \newline
3. A generated answer

If the generated answer expresses the same
meaning/topic, mark CORRECT.

Question: \promptvar{<question>} \newline
Gold answer: \promptvar{<gold answer>} \newline
Generated answer: \promptvar{<generated answer>}

Return JSON only:

\texttt{\{"label": "CORRECT"\}} \newline
or \newline
\texttt{\{"label": "WRONG"\}}
\end{promptbox}

\end{document}